# OPTIMIZATION IN SMT WITH $\mathcal{LA}(\mathbb{Q})$ COST FUNCTIONS

Roberto Sebastiani and Silvia Tomasi

January 2012

Technical Report # DISI-12-003

# Optimization in SMT with $\mathcal{LA}(\mathbb{Q})$ Cost Functions

Roberto Sebastiani and Silvia Tomasi

DISI, University of Trento, Italy

**Abstract.** In the contexts of automated reasoning and formal verification, important *decision* problems are effectively encoded into Satisfiability Modulo Theories (SMT). In the last decade efficient SMT solvers have been developed for several theories of practical interest (e.g., linear arithmetic, arrays, bit-vectors). Surprisingly, very few work has been done to extend SMT to deal with *optimization* problems; in particular, we are not aware of any work on SMT solvers able to produce solutions which minimize cost functions over *arithmetical* variables. This is unfortunate, since some problems of interest require this functionality.
In this paper we start filling this gap. We present and discuss two general procedures for leveraging SMT to handle the minimization of $\mathcal{LA}(\mathbb{Q})$ cost functions, combining SMT with standard minimization techniques. We have implemented the procedures within the MathSAT SMT solver. Due to the absence of competitors in AR and SMT domains, we have experimentally evaluated our implementation against state-of-the-art tools for the domain of *linear generalized disjunctive programming (LGDP)*, which is closest in spirit to our domain, on sets of problems which have been previously proposed as benchmarks for the latter tools. The results show that our tool is very competitive with, and often outperforms, these tools on these problems, clearly demonstrating the potential of the approach.

# Table of Contents

# 1 Introduction

In the contexts of automated reasoning (AR) and formal verification (FV), important *decision* problems are effectively encoded into and solved as Satisfiability Modulo Theories (SMT) problems. In the last decade efficient SMT solvers have been developed, that combine the power of modern conflict-driven clause-learning (CDCL) SAT solvers with dedicated decision procedures ($\mathcal{T}$-Solver*s*) for several first-order theories of practical interest like, e.g., those of linear arithmetic over the rationals ($\mathcal{LA}(\mathbb{Q})$) or the integers ($\mathcal{LA}(\mathbb{Z})$), of arrays ($\mathcal{AR}$), of bit-vectors ($\mathcal{BV}$), and their combinations. (See [36, 13] for an overview.)

Many SMT-encodable problems of interest, however, may require also the capability of finding models that are *optimal* wrt. some cost function over continuous arithmetical variables. [1] E.g., in (SMT-based) *planning with resources* [40] a plan for achieving a certain goal must be found which not only fulfills some resource constraints (e.g. on time, gasoline consumption, ...) but that also minimizes the usage of some such resource; in SMT-based *model checking with timed or hybrid systems* (e.g. [9, 8]) you may want to find executions which minimize some parameter (e.g. elapsed time), or which minimize/maximize the value of some constant parameter (e.g., a clock timeout value) while fulfilling/violating some property (e.g., minimize the closure time interval of a rail-crossing while preserving safety). This also involves, as particular subcases, problems which are traditionally addressed as *linear disjunctive programming (LDP)* [10] or *linear generalized disjunctive programming (LGDP)* [32, 35], or as SAT/SMT with Pseudo-Boolean (PB) constraints and Weighted Max-SAT/SMT problems [33, 24, 29, 17, 7]. Notice that the two latter problems can be easily encoded into each other.

Surprisingly, very few work has been done to extend SMT to deal with optimization problems [29, 17, 7]; in particular, to the best of our knowledge, all such works aim at minimizing cost functions over *Boolean* variables (i.e., SMT with PB cost functions or MAX-SMT), whilst we are not aware of any work on SMT solvers able to produce solutions which minimize cost functions over *arithmetical* variables. Notice that the former can be easily encoded into the latter, but not vice versa (see §3).

In this paper we start filling this gap. We present two general procedures for adding to SMT($\mathcal{LA}(\mathbb{Q}) \cup \mathcal{T}$) the functionality of finding models minimizing some $\mathcal{LA}(\mathbb{Q})$ cost variable —$\mathcal{T}$ being some (possibly empty) stably-infinite theory s.t. $\mathcal{T}$ and $\mathcal{LA}(\mathbb{Q})$ are signature-disjoint. These two procedures combine standard SMT and minimization techniques: the first, called *offline*, is much simpler to implement, since it uses an incremental SMT solver as a black-box, whilst the second, called *inline*, is more sophisticate and efficient, but it requires modifying the code of the SMT solver. (This distinction is important, since the source code of most SMT solvers is not publicly available.)

We have implemented these procedures within the MATHSAT5 SMT solver [5]. Due to the absence of competitors from AR and SMT domains, we have experimentally evaluated our implementation against state-of-the-art tools for the domain of LGDP, which is closest in spirit to our domain, on sets of problems which have been previously proposed as benchmarks for the latter tools. (Notice that LGDP is limited to plain $\mathcal{LA}(\mathbb{Q})$, so that, e.g., it cannot handle combination of theories like $\mathcal{LA}(\mathbb{Q}) \cup \mathcal{T}$.) The results show that our tool is very competitive with, and often outperforms, these tools on these problems, clearly demonstrating the potential of the approach.

---

[1] Although we refer to quantifier-free formulas, as it is frequent practice in SAT and SMT, with a little abuse of terminology we often call "Boolean variables" the propositional atoms and we call "variables" the Skolem constants $x_i$ in $\mathcal{LA}(\mathbb{Q})$-atoms like "$3x_1 - 2x_2 + x_3 \leq 3$".

**Related work.** The idea of optimization in SMT was first introduced by Nieuwenhuis & Oliveras [29], who presented a very-general logical framework of "SMT with progressively stronger theories" (e.g., where the theory is progressively strengthened by every new approximation of the minimum cost), and present implementations for Max-SAT/SMT based on this framework. Cimatti et al. [17] introduced the notion of "Theory of Costs" $\mathcal{C}$ to handle PB cost functions and constraints by an ad-hoc and independent "$\mathcal{C}$-solver" in the standard lazy SMT schema, and implemented a variant of MathSAT tool able to handle SAT/SMT with PB constraints and to minimize PB cost functions. The SMT solver YICES [7] also implements Max-SAT/SMT, but we are not aware of any document describing the procedures used there.

*Mixed Integer Linear Programming* (MILP) is an extension of Linear Programming (LP) involving both discrete and continuous variables. A large variety of techniques and tools for MILP are available, mostly based on efficient combinations of LP, *branch-and-bound* search mechanism and *cutting-plane* methods (see e.g. [25]). *Linear Disjunctive Programming* (LDP) problems are LP problems where linear constraints are connected by conjunctions and disjunctions [10]. Closest to our domain, *Linear Generalized Disjunctive Programming (LGDP)*, is a generalization of LDP which has been proposed in [32] as an alternative model to the MILP problem. Unlike MILP, which is based entirely on algebraic equations and inequalities, the LGDP model allows for combining algebraic and logical equations with Boolean propositions through Boolean operations, providing a much more natural representation of discrete decisions. (To this extent, LGDP and OMT($\mathcal{LA}(\mathbb{Q})$), without the "$\cup \mathcal{T}$", can be encoded into each other.) Current approaches successfully address LGDP by reformulating and solving it as a MILP problem [32, 39, 34, 35]; these reformulations focus on efficiently encoding disjunctions and logic propositions into MILP, so that to be fed to an efficient MILP solver like CPLEX.

**Content.** The rest of the paper is organized as follows: in §2 we provide some background knowledge about MILP, LGDP, SAT and SMT; in §3 we define the problem addressed, and show how it generalizes many known optimization problems; in §4 we present our novel procedures; in §5 we present an experimental evaluation; in §6 we briefly conclude and highlight directions for future work.

## 2 Background

In this section we provide the necessary background about SAT (§2.2) and SMT (§2.3). We assume a basic background knowledge about operational research and logic. We provide a uniform notation for SAT and SMT: we use boldface lowcase letters $\mathbf{a}, \mathbf{y}$ for arrays and boldface upcase letters $\mathbf{A}, \mathbf{Y}$ for matrices, standard lowcase letters $a, y$ for single rational variables/constants or indices and standard upcase letters $A, Y$ for Boolean atoms and index sets; we use the first five letters in the various forms $\mathbf{a}, ...\mathbf{e}$, $... A, ...E$, to denote *constant* values, the last five $\mathbf{v}, ...\mathbf{z}, ... V, ...Z$ to denote *variables*, and the letters $i, j, k, I, J, K$ for indexes and index sets respectively, pedices $._j$ denote the $j$-th element of an array or matrix, whilst apices $.^{ij}$ are just indexes, being part of the name of the element. We use lowcase Greek letters $\varphi, \phi, \psi, \mu, \eta$ for denoting formulas and upcase ones $\Phi, \Psi$ for denoting sets of formulas.

We assume the standard syntactic and semantic notions of propositional logic. Given a non-empty set of primitive propositions $\mathcal{P} = \{p_1, p_2, \ldots\}$, the language of propositional logic is the least set of formulas containing $\mathcal{P}$ and the primitive constants $\top$ and $\bot$ ("true" and "false") and closed under the set of standard propositional connectives $\{\neg, \wedge, \vee, \rightarrow, \leftrightarrow\}$. We call a *propositional atom* every primitive proposition in $\mathcal{P}$, and a *propositional literal* every propositional atom (*positive literal*) or its negation (*negative literal*). We implicitly remove double negations: e.g., if $l$ is the negative literal $\neg p_i$, by $\neg l$ we mean $p_i$ rather than $\neg\neg p_i$. We represent a truth assignment $\mu$ as a conjunction of literals $\bigwedge_i l_i$ (or indifferently as a set of literals $\{l_i\}_i$) with the intended meaning that a positive [resp. negative] literal $p_i$ means that $p_i$ is assigned to true [resp. false], and we see $\neg\mu$ as the clause $\bigvee_i \neg l_i$

A propositional formula is in *conjunctive normal form, CNF,* if it is written as a conjunction of disjunctions of literals: $\bigwedge_i \bigvee_j l_{ij}$. Each disjunction of literals $\bigvee_j l_{ij}$ is called a *clause*. A *unit clause* is a clause with only one literal.

### 2.1 Linear Generalized Disjunctive Programming

*Mixed Integer Linear Programming* (MILP) is an extension of Linear Programming (LP) involving both discrete and continuous variables [25]. MILP problems have the following form:
$$\min\{\mathbf{cx} : \mathbf{Ax} \geq \mathbf{b}, \mathbf{x} \geq 0, \mathbf{x}_j \in \mathbb{Z} \,\forall j \in I\} \tag{1}$$
where $\mathbf{A}$ is a matrix, $\mathbf{c}$ and $\mathbf{b}$ are constant vectors and $\mathbf{x}$ the variable vector. They are effectively solved by integrating the *branch-and-bound* search mechanism and *cutting plane* methods, resulting in a *branch-and-cut* approach. The branch-and-bound search iteratively partitions the solution space of the original MILP problem into subproblems and solves their LP relaxation (i.e. a MILP problem where the integrality constraint on the variables $\mathbf{x}_j$, for all $j \in I$, is dropped) until all variables are integral in the LP relaxation. The solutions that are infeasible in the original problem guide the search in the following way. When the optimal solution of a LP relaxation is greater than or equal to the optimal solution found so far, the search backtracks since there cannot exist a better solution. Otherwise, if a variable $\mathbf{x}_j$ is required to be integral in the original problem, the rounding of its value $a$ in the LP relaxation suggests how to branch by requiring $\mathbf{x}_j \leq \lfloor a \rfloor$ in one branch and $\mathbf{x}_j \geq \lfloor a \rfloor + 1$ in the other. Cutting planes (e.g. Gomory mixed-integer and lift-and-project cuts [25]) are linear inequalities that can be inferred and added to the original MILP problem and its subproblems in order to cut away non-integer solutions of the LP relaxation and obtain a tighter relaxation.

*Linear Disjunctive Programming* (LDP) problems are LP problems where linear constraints are connected by the logical operations of conjunction and disjunction [10]. Typically, the constraint set is expressed by a disjunction of linear systems:
$$\bigvee_{i \in I} (\mathbf{A}^i \mathbf{x} \geq \mathbf{b}^i) \tag{2}$$

or, alternatively, as:

$$(\mathbf{A}\mathbf{x} \geq \mathbf{b}) \wedge \bigwedge_{j=1}^{t} \bigvee_{k \in I_j} (\mathbf{c}^k \mathbf{x} \geq d^k) \tag{3}$$

where $\mathbf{A}\mathbf{x} \geq \mathbf{b}$ consists of the inequalities common to $\mathbf{A}^i \mathbf{x} \geq \mathbf{b}^i$ for $i \in I$, $I_j$ for $j = 1, \ldots, t$ contains one inequality of each system $\mathbf{A}^i \mathbf{x} \geq \mathbf{b}^i$ and $t$ is the number of sets $I_j$ having this property. LDP problems are effectively solved by the lift-and-project approach which combines a family of cutting planes, called lift-and-project cuts, and the branch-and-bound schema (see, e.g., [11]).

*Linear Generalized Disjunctive Programming* (LGDP) extends LDP by combining algebraic and logical equations through disjunctions and logic propositions. The general formulation of a LGDP problem is the following [32]:

$$\begin{aligned}
\min \quad & \sum_{\forall k \in K} \mathbf{z}_k + \mathbf{dx} \\
\text{s.t.} \quad & \mathbf{Bx} \leq \mathbf{b} \\
& \bigvee_{j \in J_k} \begin{bmatrix} Y^{jk} \\ \mathbf{A}^{jk}\mathbf{x} \geq \mathbf{a}^{jk} \\ \mathbf{z}_k = c^{jk} \end{bmatrix} \quad \forall k \in K \\
& \phi \\
& 0 \leq \mathbf{x} \leq \mathbf{e} \\
& \mathbf{z}_k \in \mathbb{R}^1_+, Y^{jk} \in \{True, False\} \; \forall j \in J_k, \forall k \in K
\end{aligned} \tag{4}$$

where $\mathbf{x}$ is a vector of rational variables, $\mathbf{z}$ is a vector representing the cost assigned to each disjunction and $c^{jk}$ are fixed charges, $\mathbf{e}$ is a vector of upper bounds for $\mathbf{x}$ and $Y^{jk}$ are Boolean variables. Each disjunction $k \in K$ is composed by two or more disjuncts $j \in J_k$ that contain a set of linear constraints $\mathbf{A}^{jk}\mathbf{x} \geq \mathbf{a}^{jk}$, where $(\mathbf{A}^{jk}, \mathbf{a}^{jk})$ is a $m_{jk} \times (n+1)$ matrix, for all $j \in J_k$ and $k \in K$, that are connected by the logical OR operator. Boolean variables $Y^{jk}$ and logic propositions $\phi$ in terms of $Y^{jk}$ (expressed in in Conjunctive Normal Form) represents discrete decisions. Only the constraints inside disjunct $j \in J_k$, where $Y^{jk}$ is true, are enforced. $\mathbf{Bx} \leq \mathbf{b}$, where $(\mathbf{B}, \mathbf{b})$ is a $m \times (n+1)$ matrix, are constraints that must hold regardless of disjuncts.

LGDP problems can be solved using MILP solvers by reformulating the original problem in different ways, big-M (BM) and convex hull (CH) are the two most common reformulations. In BM Boolean variables $Y^{jk}$ and logic constraints $\phi$ are replaced by binary variables $\mathbf{Y}_{jk}$ and linear inequalities as follows [32]:

$$\begin{aligned}
\min \quad & \sum_{\forall k \in K} \sum_{\forall j \in J_k} c^{jk} \mathbf{Y}_{jk} + \mathbf{dx} \\
\text{s.t.} \quad & \mathbf{Bx} \leq \mathbf{b} \\
& \mathbf{A}^{jk}\mathbf{x} - \mathbf{a}^{jk} \leq \mathbf{M}^{jk}(1 - \mathbf{Y}_{jk}) \; \forall j \in J_k, \forall k \in K \\
& \sum_{\forall j \in J_k} \mathbf{Y}_{jk} = 1 \quad \forall k \in K \\
& \mathbf{DY} \leq \mathbf{D}' \\
& \mathbf{x} \in \mathbb{R}^n_+, \mathbf{Y}_{jk} \in \{0, 1\} \quad \forall j \in J_k, \forall k \in K
\end{aligned} \tag{5}$$

where $\mathbf{M}^{jk}$ are the '"big-M" parameters that makes redundant the system of constraint $j \in J_k$ in the disjunction $k \in K$ when $\mathbf{Y}_{jk} = 0$ and the constraints $\mathbf{DY} \leq \mathbf{D}'$ are derived from $\phi$.

In CH Boolean variables $Y^{jk}$ are replaced by binary variables $\mathbf{Y}_{jk}$ and the variables $\mathbf{x} \in \mathbb{R}^n$ are disaggregated into new variables $\mathbf{v} \in \mathbb{R}^n$ in the following way:

$$
\begin{aligned}
\min \quad & \sum_{\forall k \in K} \sum_{\forall j \in J_k} c^{jk} \mathbf{Y}_{jk} + \mathbf{dx} \\
\text{s.t.} \quad & \mathbf{Bx} \leq \mathbf{b} \\
& \mathbf{A}^{kj} \mathbf{v}^{jk} \leq \mathbf{a}^{jk} \mathbf{Y}_{jk} && \forall j \in J_k, \forall k \in K \\
& \mathbf{x} = \sum_{\forall j \in J_k} \mathbf{v}_{jk} && \forall k \in K \\
& \mathbf{v}_{jk} \leq \mathbf{Y}_{jk} \mathbf{e}^{jk} && \forall j \in J_k, \forall k \in K \\
& \sum_{\forall j \in J_k} \mathbf{Y}_{jk} = 1 && \forall k \in K \\
& \mathbf{DY} \leq \mathbf{D}' \\
& \mathbf{x}, \mathbf{v} \in \mathbb{R}^n_+, \mathbf{Y}_{jk} \in \{0,1\} && \forall j \in J_k, \forall k \in K
\end{aligned}
\quad (6)
$$

where constant $\mathbf{e}^{jk}$ are upper bounds for variables $\mathbf{v}$ chosen to match the upper bounds on the variables $\mathbf{x}$.

Sawaya and Grossman [34] observed two facts. First, the relaxation of BM is often weak causing a higher number of nodes examined in the branch-and-bound search. Second, the disaggregated variables and new constraints increase the size of the reformulation leading to a high computational effort. In order to overcome these issues, they proposed a cutting plane method that consists in solving a sequence of BM relaxations with cutting planes that are obtained from CH relaxations. They provided an evaluation of the presented algorithm on three different problems: strip-packing, retrofit planning and zero-wait job-shop scheduling problems.

## 2.2 SAT and CDCL SAT Solvers

We present here a brief description on how a Conflict-Driven Clause-Learning (CDCL) SAT solver works. (We refer the reader, e.g., to [27] for a detailed description.)

We assume the input propositional formula $\varphi$ is in CNF. (If not, it is first CNF-ized as in [31].) The assignment $\mu$ is initially empty, and it is updated in a stack-based manner. The SAT solver performs an external loop, alternating three main phases: *Decision*, *Boolean Constraint Propagation (BCP)* and *Backjumping and Learning*.

During *Decision* an unassigned literal $l$ from $\varphi$ is selected according to some heuristic criterion, and it is pushed into $\mu$. ($l$ is called *decision literal* and the number of decision literals in $\mu$ after this operation is called the *decision level* of $l$.)

Then *BCP* iteratively deduces literals $l$ deriving from the current assignment and pushes them into $\mu$. *BCP* is based on the iterative application of *unit propagation*: if all but one literals in a clause $\psi$ are false, then the lonely unassigned literal $l$ is added to $\mu$, all negative occurrences of $l$ in other clauses are declared false and all clauses with positive occurrences of $l$ are declared satisfied. Current SAT solvers include rocket-fast implementations of *BCP* based on the *two-watched-literal scheme*, see [27]. *BCP* is repeated until either no more literals can be deduced, so that the loop goes back to another Decision step, or no more Boolean variable can be assigned, so that the SAT solver ends returning SAT, or $\mu$ falsifies some clause $\psi$ of $\varphi$ (*conflicting clause*).

In the latter case, *Backjumping and learning* are performed. A process of *conflict analysis* [2] detects a subset $\eta$ of $\mu$ which actually caused the falsification of $\psi$ (*conflict set*) [3] and the decision level `blevel` where to backtrack. Otherwise, the *conflict clause* $\psi' \stackrel{\text{def}}{=} \neg \eta$ is added to $\varphi$ (*learning*) and the procedure backtracks up to blevel (*backjumping*), popping out of $\mu$ all literals whose decision level is greater than blevel. When two contradictory literals $l, \neg l$ are assigned at level 0, the loop terminates, returning UNSAT.

Notice that CDCL SAT solvers implement "safe" strategy for discharging clauses when no more necessary, which guarantee the use of polynomial space without affecting the termination, correctness and completeness of the procedure. (See e.g. [27, 30].)

Many modern CDCL SAT solvers provide a *stack-based incremental interface* (see e.g. [21]), by which it is possible to push/pop sub-formulas $\phi_i$ into a stack of formulas $\Phi \stackrel{\text{def}}{=} \{\phi_1, ..., \phi_k\}$, and check incrementally the satisfiability of $\bigwedge_{i=1}^{k} \phi_i$. The interface maintains the *status* of the search from one call to the other, in particular it records the learned clauses (plus other information). Consequently, when invoked on $\Phi$ the solver can reuse a clause $\psi$ which was learned during a previous call on some $\Phi'$ if $\psi$ was derived only from clauses which are still in $\Phi$; [4] in particular, if $\Phi' \subseteq \Phi$, then the solver can reuse all clauses learned while solving $\Phi'$. Another important feature of many incremental CDCL SAT solvers is the capability, when $\Phi$ is found unsatisfiable, of returning the subset of formulas in $\Phi$ which caused the unsatisfiability of $\Phi$. (This is a subcase of the more general problem of finding the *unsatisfiable core* of a formula, see e.g. [26].) Notice that such subset is not unique, and it is not necessarily minimal.

### 2.3 SMT and Lazy SMT Solvers

A *theory solver for* $\mathcal{T}$, $\mathcal{T}$-Solver, is a procedure able to decide the $\mathcal{T}$-satisfiability of a conjunction/set $\mu$ of $\mathcal{T}$-literals. If $\mu$ is $\mathcal{T}$-unsatisfiable, then $\mathcal{T}$-Solver returns UNSAT and the set/conjunction $\eta$ of $\mathcal{T}$-literals in $\mu$ which was found $\mathcal{T}$-unsatisfiable; $\eta$ is called a $\mathcal{T}$-*conflict set*, and $\neg \eta$ a $\mathcal{T}$-*conflict clause*. If $\mu$ is $\mathcal{T}$-satisfiable, then $\mathcal{T}$-Solver returns SAT; it may also be able to return some unassigned $\mathcal{T}$-literal $l \notin \mu$[5] s.t. $\{l_1, ..., l_n\} \models_\mathcal{T} l$, where $\{l_1, ..., l_n\} \subseteq \mu$. We call this process $\mathcal{T}$-*deduction* and $(\bigvee_{i=1}^{n} \neg l_i \vee l)$ a $\mathcal{T}$-*deduction clause*. Notice that $\mathcal{T}$-conflict and $\mathcal{T}$-deduction clauses are valid in $\mathcal{T}$. We call them $\mathcal{T}$-*lemmas*.

Given a $\mathcal{T}$-formula $\varphi$, the formula $\varphi^p$ obtained by rewriting each $\mathcal{T}$-atom in $\varphi$ into a fresh atomic proposition is the *Boolean abstraction* of $\varphi$, and $\varphi$ is the *refinement* of $\varphi^p$. Notationally, we indicate by $\varphi^p$ and $\mu^p$ the Boolean abstraction of $\varphi$ and $\mu$, and by $\varphi$ and $\mu$ the refinements of $\varphi^p$ and $\mu^p$ respectively. With a little abuse of notation, we say that $\mu^p$ is $\mathcal{T}$-(un)satisfiable iff $\mu$ is $\mathcal{T}$-(un)satisfiable.

---

[2] When a clause $\psi$ is falsified by the current assignment a *conflict clause* $\psi'$ is computed from $\psi$ s.t. $\psi'$ contains only one literal $l_u$ which has been assigned at the last decision level. $\psi'$ is computed starting from $\psi' = \psi$ by iteratively resolving $\psi'$ with the clause $\psi_l$ causing the unit-propagation of some literal $l$ in $\psi'$ until some stop criterion is met.

[3] that is, $\eta$ is enough to force the unit-propagation of the literals causing the failure of $\psi$.

[4] provided $\psi$ was not discharged in the meantime.

[5] taken from a set of all the available $\mathcal{T}$-literals; when combined with a SAT solver, such set would be the set of all the $\mathcal{T}$-literals occurring in the input formula to solve.

We say that the truth assignment $\mu$ *propositionally satisfies* the formula $\varphi$, written $\mu \models_p \varphi$, if $\mu^p \models \varphi^p$.

In a lazy SMT($\mathcal{T}$) solver, the Boolean abstraction $\varphi^p$ of the input formula $\varphi$ is given as input to a CDCL SAT solver, and whenever a satisfying assignment $\mu^p$ is found s.t. $\mu^p \models \varphi^p$, the corresponding set of $\mathcal{T}$-literals $\mu$ is fed to the $\mathcal{T}$-Solver; if $\mu$ is found $\mathcal{T}$-consistent, then $\varphi$ is $\mathcal{T}$-consistent; otherwise, $\mathcal{T}$-Solver returns the $\mathcal{T}$-conflict set $\eta$ causing the inconsistency, so that the clause $\neg \eta^p$ is used to drive the backjumping and learning mechanism of the SAT solver. The process proceeds until either a $\mathcal{T}$-consistent assignment $\mu$ is found, or no more assignments are available ($\varphi$ is $\mathcal{T}$-inconsistent).

Important optimizations are *early pruning* and $\mathcal{T}$-*propagation*: the $\mathcal{T}$-Solver is invoked also when an assignment $\mu$ is still under construction: if it is $\mathcal{T}$-unsatisfiable, then the procedure backtracks, without exploring the (up to exponentially-many) extensions of $\mu$; if not, and if the $\mathcal{T}$-Solver is able to perform a $\mathcal{T}$-deduction $\{l_1, ..., l_n\} \models_{\mathcal{T}} l$, then $l$ can be unit-propagated, and the $\mathcal{T}$-deduction clause $(\bigvee_{i=1}^n \neg l_i \vee l)$ can be used in backjumping and learning.

Another optimization is *pure-literal filtering*: if some $\mathcal{LA}(\mathbb{Q})$-atoms occur only positively [resp. negatively] in the original formula (learned clauses are ignored), then we can safely drop every negative [resp. positive] occurrence of them from the assignment $\mu$ to be checked by the $\mathcal{T}$-solver [36]. (Intuitively, since such occurrences play no role in satisfying the formula, the resulting partial assignment $\mu^{p'}$ still satisfies $\varphi^p$.) The benefits of this action is twofold: (i) reduces the workload for the $\mathcal{T}$-Solver by feeding it smaller sets; (ii) increases the chance of finding a $\mathcal{T}$-consistent satisfying assignment by removing "useless" $\mathcal{T}$-literals which may cause the $\mathcal{T}$-inconsistency of $\mu$.

The above schema is a coarse abstraction of the procedures underlying all the state-of-the-art lazy SMT tools. (The interested reader is pointed to, e.g., [30, 36, 13] for details and further references.) Importantly, some SMT solvers, including MATHSAT, inherit from their embedded SAT solver the capabilities of working incrementally and of returning the subset of input formulas causing the inconsistency, as described in §2.2.

The *Theory of Linear Arithmetic on the rationals* ($\mathcal{LA}(\mathbb{Q})$) and on the integer ($\mathcal{LA}(\mathbb{Z})$) is one of the theories of main interest in SMT. It is a first-order theory whose atoms are of the form $(a_1 x_1 + \ldots + a_n x_n \diamond b)$ (i.e. $(\mathbf{a}\mathbf{x} \diamond b)$) s.t $\diamond \in \{=, \neq, <, >, \leq, \geq,\}$. *Difference logic* on $\mathbb{Q}$ ($\mathcal{DL}(\mathbb{Q})$) is an important sub-theory of $\mathcal{LA}(\mathbb{Q})$, in which all atoms are in the form $(x_1 - x_2 \diamond b)$.

Efficient incremental and backtrackable procedures have been conceived in order to decide $\mathcal{LA}(\mathbb{Q})$ [20], $\mathcal{LA}(\mathbb{Z})$ [22] and $\mathcal{DL}$ [19]. In particular, for $\mathcal{LA}(\mathbb{Q})$ substantially all SMT solvers implement variants of the simplex-based algorithm by Dutertre and de Moura [20] which is specifically designed for integration in a lazy SMT solver, since it is fully incremental and backtrackable and allows for aggressive $\mathcal{T}$-deduction. Another benefit of such algorithm is that it handles strict inequalities directly. Its method is based on the fact that a set of $\mathcal{LA}(\mathbb{Q})$ atoms $\Gamma$ containing strict inequalities $S = \{0 < t_1, \ldots, 0 < t_n\}$ is satisfiable iff there exists a rational number $\epsilon > 0$ such that $\Gamma_\epsilon \stackrel{\text{def}}{=} (\Gamma \cup S_\epsilon) \setminus S$ is satisfiable, s.t. $S_\epsilon \stackrel{\text{def}}{=} \{\epsilon \leq t_1, \ldots, \epsilon \leq t_n\}$. The idea of [20] is that of treating the *infinitesimal parameter* $\epsilon$ symbolically instead of explicitly computing its value. Strict bounds ($x < b$) are replaced with weak ones ($x \leq b - \epsilon$), and the operations on bounds are adjusted to take $\epsilon$ into account. (We refer the reader to [20] for details.)

## 3   Optimization in SMT($\mathcal{LA}(\mathbb{Q}) \cup \mathcal{T}$)

Let $\mathcal{T}$ be some stably infinite theory with equality s.t. $\mathcal{LA}(\mathbb{Q})$ and $\mathcal{T}$ are signature-disjoint, as in [28]. ($\mathcal{T}$ can be itself a combination of theories.) We call an *Optimization Modulo $\mathcal{LA}(\mathbb{Q}) \cup \mathcal{T}$ problem, OMT($\mathcal{LA}(\mathbb{Q}) \cup \mathcal{T}$)*, a pair $\langle \varphi, \mathsf{cost} \rangle$ such that $\varphi$ is a SMT($\mathcal{LA}(\mathbb{Q}) \cup \mathcal{T}$) formula and cost is a $\mathcal{LA}(\mathbb{Q})$ variable occurring in $\varphi$, representing the cost to be minimized. The problem consists in finding a model $\mathcal{M}$ for $\varphi$ (if any) whose value of cost is minimum. We call an *Optimization Modulo $\mathcal{LA}(\mathbb{Q})$ problem (OMT($\mathcal{LA}(\mathbb{Q})$))* an SMT($\mathcal{LA}(\mathbb{Q}) \cup \mathcal{T}$) problem where $\mathcal{T}$ is empty. If $\varphi$ is in the form $\varphi' \wedge (\mathsf{cost} < c)$ [resp. $\varphi' \wedge \neg(\mathsf{cost} < c)$] for some value $c \in \mathbb{Q}$, then we call $c$ an *upper bound* [resp. *lower bound*] for cost. If ub [resp lb ] is the minimum upper bound [resp. the maximum lower bound] for $\varphi$, we also call the interval [lb, ub[ the *range* of cost. [6]

These definitions capture many interesting optimizations problems. LP is a particular subcase of OMT($\mathcal{LA}(\mathbb{Q})$) with no Boolean component, such that $\varphi \stackrel{\text{def}}{=} \varphi' \wedge (\mathsf{cost} = \sum_i \mathbf{a}_i \mathbf{x}_i)$ and $\varphi' = \bigwedge_j (\sum_i \mathbf{A}_{ij} \mathbf{x}_i \leq \mathbf{b}_j)$. LDP is easily encoded into OMT($\mathcal{LA}(\mathbb{Q})$), since (2) and (3) can be written as

$$\bigvee_i \bigwedge_j (\mathbf{A}_j^i \mathbf{x} \geq \mathbf{b}_j^i) \text{ or } \bigwedge_j (\mathbf{A}_j \mathbf{x} \geq \mathbf{b}_j) \wedge \bigwedge_{j=1}^t \bigvee_{k \in I_j} (\mathbf{c}^k \mathbf{x} \geq d^k) \qquad (7)$$

respectively, where $\mathbf{A}_j^i$ and $\mathbf{A}_j$ are respectively the $j$th row of the matrices $\mathbf{A}^i$ and $\mathbf{A}$, $\mathbf{b}_j^i$ and $\mathbf{b}_j$ are respectively the $j$th row of the vectors $\mathbf{b}^i$ and $\mathbf{b}$. Since the left equation (7) is not in CNF, the CNF-ization process of [31] is then applied.

LGDP (4) is straightforwardly encoded into a OMT($\mathcal{LA}(\mathbb{Q})$) problem $\langle \varphi, \mathsf{cost} \rangle$:

$$\begin{aligned} \varphi \stackrel{\text{def}}{=} & (\mathsf{cost} = \sum_{\forall k \in K} \mathbf{z}_k + \mathbf{dx}) \wedge [[\mathbf{Bx} \leq \mathbf{b}]] \wedge \phi \wedge [[\mathbf{0} \leq \mathbf{x}]] \wedge [[\mathbf{x} \leq \mathbf{e}]] \\ & \wedge \bigwedge_{k \in K} \bigvee_{j \in J_k} (Y^{jk} \wedge [[\mathbf{A}^{jk} \mathbf{x} \geq \mathbf{a}^{jk}]] \wedge (\mathbf{z}_k = c^{jk})) \end{aligned} \qquad (8)$$

s.t. $[[\mathbf{x} \bowtie \mathbf{a}]]$ and $[[\mathbf{Ax} \bowtie \mathbf{a}]]$ are abbreviations respectively for $\bigwedge_i (\mathbf{x}_i \bowtie \mathbf{a}_i)$ and $\bigwedge_i (\mathbf{A}_i . \mathbf{x} \bowtie \mathbf{a}_i)$, $\bowtie \in \{=, \neq \leq, \geq, <, >\}$. Since (8) is not in CNF, the CNF-ization process of [31] is then applied.

Pseudo-Boolean (PB) constraints (see [33]) in the form ($\sum_i \mathbf{a}_i X^i \leq b$), s.t. $X^i$ are Boolean atoms and $\mathbf{a}_i$ constant values in $\mathbb{Q}$, and cost functions $\mathsf{cost} = \sum_i \mathbf{a}_i X^i$, are encoded into OMT($\mathcal{LA}(\mathbb{Q})$) by rewriting each PB-term $\sum_i \mathbf{a}_i X^i$ into the $\mathcal{LA}(\mathbb{Q})$-term $\sum_i \mathbf{x}_i$, $\mathbf{x}$ being an array of fresh $\mathcal{LA}(\mathbb{Q})$ variables, and by conjoining to $\varphi$ the formula:

$$\bigwedge_i ((\neg X^i \vee (\mathbf{x}_i = \mathbf{a}_i)) \wedge (X^i \vee (\mathbf{x}_i = 0)) \wedge (\mathbf{x}_i \geq 0) \wedge (\mathbf{x}_i \leq \mathbf{a}_i)^{[7]}). \qquad (9)$$

---

[6] We adopt the convention of defining upper bounds to be strict and lower bounds to be non-strict for a practical reason: typically an upper bound ($\mathsf{cost} < c$) derives from the fact that a model $\mathcal{M}$ of cost $c$ has been previously found, whilst a lower bound $\neg(\mathsf{cost} < c)$ derives either from the user's knowledge (e.g. "the cost cannot be lower than zero") of from the fact that the formula $\varphi \wedge (\mathsf{cost} < c)$ has been previously found $\mathcal{T}$-unsatisfiable whilst $\varphi$ is not.

[7] The term "$(\mathbf{x}_i \geq 0) \wedge (\mathbf{x}_i \leq \mathbf{a}_i)$" is not necessary, but it improves the performances of the SMT($\mathcal{LA}(\mathbb{Q})$) solver because it allows for exploiting the early-pruning technique.

Moreover, since Max-SAT (see [24]) [resp. Max-SMT (see [29, 17, 7])] can be encoded into SAT [resp. SMT] with PB constraints (see e.g. [29, 17]), then optimization problems for SAT with PB constraints and Max-SAT can be encoded into OMT($\mathcal{LA}(\mathbb{Q})$), whilst those for SMT($\mathcal{T}$) with PB constraints and Max-SMT can be encoded into OMT($\mathcal{LA}(\mathbb{Q}) \cup \mathcal{T}$) (assuming $\mathcal{T}$ matches the definition above).

We remark the deep difference between OMT($\mathcal{LA}(\mathbb{Q})$)/OMT($\mathcal{LA}(\mathbb{Q}) \cup \mathcal{T}$) and the problem of SAT/SMT with PB constraints and cost functions (or Max-SAT/SMT) addressed in [29, 17]. With the latter problem, the cost is a deterministic consequence of a truth assignment to the atoms of the formula, so that the search has only a Boolean component, consisting in finding the cheapest truth assignment. With OMT($\mathcal{LA}(\mathbb{Q})$)/ OMT($\mathcal{LA}(\mathbb{Q}) \cup \mathcal{T}$), instead, for every satisfying assignment $\mu$ it is also necessary to find the minimum-cost $\mathcal{LA}(\mathbb{Q})$-model for $\mu$, so that the search has both a Boolean and a $\mathcal{LA}(\mathbb{Q})$-component.

## 4 Procedures for OMT($\mathcal{LA}(\mathbb{Q})$) and OMT($\mathcal{LA}(\mathbb{Q}) \cup \mathcal{T}$)

It may be noticed that very naive OMT($\mathcal{LA}(\mathbb{Q})$) or OMT($\mathcal{LA}(\mathbb{Q}) \cup \mathcal{T}$) procedures could be straightforwardly implemented by performing a sequence of calls to an SMT solver on formulas like $\varphi \wedge (\mathsf{cost} \geq \mathsf{l}_i) \wedge (\mathsf{cost} < \mathsf{u}_i)$, each time restricting the range $[\mathsf{l}_i, \mathsf{u}_i[$ according to a linear-search of binary-search schema. With the former schema, every time the SMT solver returns a model of cost $c_i$, a new constraint $(\mathsf{cost} < c_i)$ would be added to $\varphi$, and the solver would be invoked again; however, the SMT solver would repeatedly generate the same $\mathcal{LA}(\mathbb{Q})$-satisfiable truth assignments, each time finding a cheaper model for it. With the latter schema the efficiency should improve; however, an initial lower-bound should be necessarily required as input (which is not the case, e.g., of the problems in §5.2.)

In this section we present more sophisticate procedures, based on the combination of SMT and minimization techniques. We first present and discuss an *offline* schema (§4.1) and an *inline* (§4.2) schema for an OMT($\mathcal{LA}(\mathbb{Q})$) procedure; then we show how to extend them to the OMT($\mathcal{LA}(\mathbb{Q}) \cup \mathcal{T}$) case (§4.2).

### 4.1 An Offline Schema for OMT($\mathcal{LA}(\mathbb{Q})$)

The general schema for the offline OMT($\mathcal{LA}(\mathbb{Q})$) procedure is displayed in Algorithm 1. It takes as input an instance of the OMT($\mathcal{LA}(\mathbb{Q})$) problem, plus optionally values for lb and ub (which are implicitly considered to be $-\infty$ and $+\infty$ if not present), and returns the model $\mathcal{M}$ of minimum cost and its cost u (the value ub if $\varphi$ is $\mathcal{LA}(\mathbb{Q})$-inconsistent). Notice that by providing a lower bound lb [resp. an upper bound ub ] the user implicitly assumes the responsibility of asserting there is no model whose cost is lower than lb [there is a model whose cost is ub ]. We represent the $\varphi$ as a set of clauses, which may be pushed or popped from the input formula-stack of an incremental SMT solver.

First, the variables l, u (defining the current range) are initialized to lb and ub respectively, the atom PIV to $\top$, and $\mathcal{M}$ is initialized to be an empty model. Then the procedure adds to $\varphi$ the bound constraints, if present, which restrict the search within

**Algorithm 1** Offline OMT($\mathcal{LA}(\mathbb{Q})$) Procedure based on Mixed Linear/Binary Search.

**Require:** $\langle \varphi, \text{cost}, \text{lb}, \text{ub} \rangle$ {ub can be $+\infty$, lb can be $-\infty$}
1: $\text{l} \leftarrow \text{lb}; \text{u} \leftarrow \text{ub}; \text{PIV} \leftarrow \top; \mathcal{M} \leftarrow \emptyset$
2: $\varphi \leftarrow \varphi \cup \{\neg(\text{cost} < \text{l}), (\text{cost} < \text{u})\}$
3: **while** ($\text{l} < \text{u}$) **do**
4:     **if** (BinSearchMode()) **then** {Binary-search Mode}
5:         pivot $\leftarrow$ ComputePivot(l, u)
6:         PIV $\leftarrow$ (cost $<$ pivot)
7:         $\varphi \leftarrow \varphi \cup \{\text{PIV}\}$
8:         $\langle \text{res}, \mu \rangle \leftarrow$ SMT.IncrementalSolve($\varphi$)
9:         $\eta \leftarrow$ SMT.ExtractUnsatCore($\varphi$)
10:     **else** {Linear-search Mode}
11:         $\langle \text{res}, \mu \rangle \leftarrow$ SMT.IncrementalSolve($\varphi$)
12:         $\eta \leftarrow \emptyset$
13:     **end if**
14:     **if** (res = SAT) **then**
15:         $\langle \mathcal{M}, \text{u} \rangle \leftarrow$ Minimize(cost, $\mu$)
16:         $\varphi \leftarrow \varphi \cup \{(\text{cost} < \text{u})\}$
17:     **else** {res = UNSAT }
18:         **if** (PIV $\notin \eta$) **then**
19:             l $\leftarrow$ u
20:         **else**
21:             l $\leftarrow$ pivot
22:             $\varphi \leftarrow \varphi \setminus \{\text{PIV}\}$
23:             $\varphi \leftarrow \varphi \cup \{\neg\text{PIV}\}$
24:         **end if**
25:     **end if**
26: **end while**
27: **return** $\langle \mathcal{M}, \text{u} \rangle$

the range [l, u[ (row 2).[8] The solution space is then explored iteratively (rows 3-26), reducing at at each loop the current range [l, u[ to explore, until the range is empty. Then $\langle \mathcal{M}, \text{u} \rangle$ is returned —$\langle \emptyset, \text{ub} \rangle$ if there is no solution in [lb, ub[— $\mathcal{M}$ being the model of minimum cost u. Each loop may work in either *linear-search* or *binary-search* mode, driven by the heuristic BinSearchMode(). Notice that if u = $+\infty$ or l = $-\infty$, then BinSearchMode() returns false.

In **linear-search mode**, steps 4-9 and 21-23 are not executed. First, an incremental SMT($\mathcal{LA}(\mathbb{Q})$) solver is invoked on $\varphi$ (row 11). (Notice that, given the incrementality of the solver, every operation in the form "$\varphi \leftarrow \varphi \cup \{\phi_i\}$" [resp. $\varphi \leftarrow \varphi \setminus \{\phi_i\}$] is implemented as a "push" [resp. "pop"] operation on the stack representation of $\varphi$, see §2.2; it is also very important to recall that during the SMT call $\varphi$ is updated with the clauses which are learned during the SMT search.) $\eta$ is set to be empty, which forces condition 18 to hold. If $\varphi$ is $\mathcal{LA}(\mathbb{Q})$-satisfiable, then it is returned res =SAT and a $\mathcal{LA}(\mathbb{Q})$-satisfiable truth assignments $\mu$ for $\varphi$. Thus Minimize is invoked on (the subset of $\mathcal{LA}(\mathbb{Q})$-literals of) $\mu$, [9] returning the model $\mathcal{M}$ for $\mu$ of minimum cost u ($-\infty$ iff the problem in unbounded). The current solution u becomes the new upper bound, thus the $\mathcal{LA}(\mathbb{Q})$-atom (cost < u) is added to $\varphi$ (row 16). Notice that if the problem is unbounded, then for some $\mu$ Minimize will return $-\infty$, forcing condition 3 to be false and the whole process to stop. If $\varphi$ is $\mathcal{LA}(\mathbb{Q})$-unsatisfiable, then no model in the current cost range [l, u[ can be found; hence the flag l is set to u, forcing the end of the loop.

---
[8] Of course literals like $\neg(\text{cost} < -\infty)$ and $(\text{cost} < +\infty)$ are not added.

In **binary-search mode** at the beginning of the loop (steps 4-9), the value pivot $\in$ ]l, u[ is computed by the heuristic function ComputePivot (in the simplest form, pivot is $(l + u)/2$), the (possibly new) atom PIV $\stackrel{\text{def}}{=}$ (cost < pivot) is pushed into the formula stack, so that to temporarily restrict the cost range to [l, pivot[; then the incremental SMT solver is invoked on $\varphi$, this time activating the feature SMT.ExtractUnsatCore, which returns also the subset $\eta$ of formulas in (the formula stack of) $\varphi$ which caused the unsatisfiability of $\varphi$ (see §2.2). This exploits techniques similar to unsat-core extraction [26]. (In practice, it suffices to say if PIV $\in \eta$.) If $\varphi$ is $\mathcal{LA}(\mathbb{Q})$-satisfiable, then the procedure behaves as in linear-search mode. If instead $\varphi$ is $\mathcal{LA}(\mathbb{Q})$-unsatisfiable, we look at $\eta$ and distinguish two subcases. If PIV does not occur in $\eta$, this means that $\varphi \setminus \{$PIV$\}$ is $\mathcal{LA}(\mathbb{Q})$-inconsistent, i.e. there is no model in the whole cost range [l, u[. Then the procedure behaves as in linear-search mode, forcing the end of the loop. Otherwise, we can only conclude that there is no model in the cost range [l, pivot[, so that we still need exploring the cost range [pivot, u[. Thus l is set to pivot, PIV is popped from $\varphi$ and its negation is pushed into $\varphi$. Then the search proceeds, investigating the cost range [pivot, u[.

We notice an important fact: if BinSearchMode() always returned true, then Algorithm 1 would not necessarily terminate. In fact, an SMT solver invoked on $\varphi$ may return a set $\eta$ containing PIV even if $\varphi \setminus$ PIV is $\mathcal{LA}(\mathbb{Q})$-inconsistent. [10] Thus, e.g., the procedure might got stuck into a "Zeno" [11] infinite loop, each time halving the cost range right-bound (e.g., $[-1, 0[, [-1/2, 0[, [-1/4, 0[,..)$. To cope with this fact, however, it suffices that BinSearchMode() returns false infinitely often, forcing then a "linear-search" call which finally detects the inconsistency. (In our implementation, we have empirically experienced the best performance with one linear-search loop after every binary-search one, because satisfiable calls are typically much cheaper then unsatisfiable ones.)

Under such hypothesis, it is straightforward to see the following facts: (i) Algorithm 1 terminates, in both modes, because there are only a finite number of candidate truth assignments $\mu$ to be enumerated, and steps 15-16 guarantee that the same assignment $\mu$ will never be returned twice by the SMT solver; (ii) it returns a model of minimum cost, because it explores the whole search space of candidate truth assignments, and for every suitable assignment $\mu$ Minimize finds the minimum-cost model for $\mu$;

---

[9] possibly after applying pure-literal filtering to $\mu$ (see §2.3).

[10] In a nutshell, a CDCL-based SMT solver implicitly builds a resolution refutation whose leaves are either clauses in $\varphi$ or $\mathcal{LA}(\mathbb{Q})$-lemmas, and the set $\eta$ represents the subset of clauses in $\varphi$ which occur as leaves of such proof (see e.g. [18] for details). If the SMT solver is invoked on $\varphi$ even $\varphi \setminus$ PIV is $\mathcal{LA}(\mathbb{Q})$-inconsistent, then it can "use" PIV and return a proof involving it even though another PIV-less proof exists.

[11] In the famous Zeno's paradox, Achilles never reaches the tortoise for a similar reason.

(iii) it requires polynomial space, under the assumption that the underlying CDCL SAT solver adopts a polynomial-size clause discharging strategy (which is typically the case of SMT solvers, including MATHSAT).

In a nutshell, Minimize is a simple extension of the simplex-based $\mathcal{LA}(\mathbb{Q})$-Solver of [20] which is invoked after one solution is found, minimizing it by standard Simplex techniques. We recall that the algorithm in [20] can handle strict inequalities. Thus, if the input problem contains strict inequalities, then Minimize temporarily treats them as non-strict ones and finds the minimum-cost solution with standard Simplex techniques. If such minimum-cost solution **x** of cost min lays only on non-strict inequalities, then **x** is a solution; otherwise, for some $\delta > 0$ and for every cost $c \in\, ]$min, min $+\, \delta]$ there exists a solution of cost $c$. (If needed, such solution is computed using the techniques for handling strict inequalities described in [20].) Thus the value min is returned, tagged as a non-strict minimum, so that the constraint (cost $\leq$ min) rather than (cost $<$ min) is added to $\varphi$.

This latter fact, however, never happens if the SMT solver (like MATHSAT) implements pure-literal filtering (§2.3) and if no strict inequalities occur in $\varphi$ with positive polarity (resp. no non-strict inequality occur in $\varphi$ with negative polarity), which is the case of most problems in this paper. In fact, due to pure-literal filtering, the only strict inequalities which may be fed to Minimize are upper-bound constraints (cost $<$ $u_i$), which by construction do not touch the minimum-cost solution **x** above.

**Discussion.** We remark a few facts about this procedure.

First, if Algorithm 1 is interrupted (e.g., by a timeout device), then u can be returned, representing the best approximation of the minimum cost found so far.

Second, the incrementality of the SMT solver (see §2.2 and §2.3) plays an essential role here, since at every call SMT.IncrementalSolve resumes the status of the search of the end of the previous call, only with tighter cost range constraints. (Notice that at each call here the solver can reuse all previously-learned clauses.) To this extent, one can see the whole process as only one SMT process, which is interrupted and resumed each time a new model is found, in which cost range constraints are progressively tightened.

Third, we notice that in Algorithm 1 all the literals constraining the cost range (i.e., ¬(cost $<$ l), (cost $<$ u)) are always added to $\varphi$ as unit clauses; thus inside SMT.IncrementalSolve these literals are immediately unit-propagated, becoming part of each truth assignment $\mu$ from the very beginning of its construction. As soon as novel $\mathcal{LA}(\mathbb{Q})$-literals are added to $\mu$ which prevent it from having a $\mathcal{LA}(\mathbb{Q})$-model of cost in $[$l, u$[$, the $\mathcal{LA}(\mathbb{Q})$-solver invoked on $\mu$ by early-pruning calls (see §2.3) returns UNSAT and the $\mathcal{LA}(\mathbb{Q})$-lemma $\neg\eta$ describing the conflict $\eta \subseteq \mu$, triggering theory-backjumping and -learning. To this extent, SMT.IncrementalSolve implicitly plays a form of *branch & bound*: (i) decide a new literal $l$ and unit- or theory-propagate the literals which derive from $l$ ("branch") and (ii) backtrack as soon as the current branch can no more be expanded into models in the current cost range ("bound"). We remark that also the constraint ¬(cost $<$ l) plays a role even in linear-search mode, since it helps pruning the search inside SMT.IncrementalSolve.

Fourth, in binary-search mode, the range-partition strategy may be even more aggressive than that of standard binary search, because the minimum cost u returned in row 15 can be significantly smaller than pivot, so that the cost range is more than halved.

Finally, unlike with other domains (e.g., search in a sorted array) the binary-search strategy here is not "obviously faster" than the linear-search one, because the unsatisfiable calls to SMT.IncrementalSolve are typically much more expensive than the satisfiable ones, because they must explore the whole Boolean search space rather than only a portion of it (although with a higher pruning power, due to the stronger constraint induced by the presence of pivot). Thus, we have a tradeoff between a typically much-smaller number of calls plus a stronger pruning power in binary search versus an average much smaller cost of the calls in linear search. To this extent, it is possible to use dynamic/adaptive strategies for ComputePivot (see [37]).

### 4.2 An Inline Schema for OMT($\mathcal{LA}(\mathbb{Q})$)

With the inline schema, the whole optimization procedure is pushed inside the SMT solver by embedding the range-minimization loop inside the CDCL Boolean-search loop of the standard lazy SMT schema of §2.3. The SMT solver, which is thus called only once, is modified as follows.

**Initialization.** The variables lb, ub, l, u, PIV, pivot, $\mathcal{M}$ are brought inside the SMT solver, and are initialized as in Algorithm 1, steps 1-2.

**Range Updating & Pivoting.** Every time the search of the CDCL SAT solver gets back to decision level 0, the range [l, u[ is updated s.t. u [resp. l ] is assigned the lowest [resp. highest] value $u_i$ [resp. $l_i$] such that the atom (cost < $u_i$) [resp. ¬(cost < $u_i$)] is currently assigned at level 0. (If u ≤ l, or two literals $l, \neg l$ are both assigned at level 0, then the procedure terminates, returning the current value of u.) Then BinSearchMode() is invoked: if it returns true, then ComputePivot computes pivot ∈ ]l, u[, and the (possibly new) atom PIV $\stackrel{\text{def}}{=}$ (cost < pivot) is decided (level 1) by the SAT solver. This mimics steps 4-7 in Algorithm 1, temporarily restricting the cost range to [l, pivot[.

**Decreasing the Upper Bound.** When an assignment $\mu$ propositionally satisfying $\varphi$ is generated which is found $\mathcal{LA}(\mathbb{Q})$-consistent by $\mathcal{LA}(\mathbb{Q})$-Solver, $\mu$ is also fed to Minimize, returning the minimum cost min of $\mu$; then the unit clause (cost < min) is learned and fed to the backjumping mechanism, which forces the SAT solver to backjump to level 0, then unit-propagating (cost < min). This case mirrors steps 14-16 in Algorithm 1, permanently restricting the cost range to [l, min[. Minimize is embedded within $\mathcal{LA}(\mathbb{Q})$-Solver, so that it is called incrementally after it, without restarting its search from scratch.

As a result of these modifications, we also have the following typical scenario (see Figure 1).

**Increasing the Lower Bound.** In binary-search mode, when a conflict occurs s.t. the conflict analysis of the SAT solver produces a conflict clause in the form ¬PIV ∨ ¬$\eta'$ s.t. all literals in $\eta'$ are assigned true at level 0 (i.e., $\varphi \wedge$ PIV is $\mathcal{LA}(\mathbb{Q})$-inconsistent), then the SAT solver backtracks to level 0, unit-propagating ¬PIV. This case mirrors steps 21-23 in Algorithm 1, permanently restricting the cost range to [pivot, u[.

Although the modified SMT solver mimics to some extent the behaviour of Algorithm 1, the "control" of the range-restriction process is handled by the standard SMT search. To this extent, notice that also other situations may allow for restricting the cost range: e.g., if $\varphi \wedge \neg$(cost < l) $\wedge$ (cost < u) $\models$ (cost $\bowtie$ m) for some atom (cost $\bowtie$ m) occurring in $\varphi$ s.t. $\mathsf{m} \in [\mathsf{l}, \mathsf{u}[$ and $\bowtie \in \{\leq, <, \geq, >\}$, then the SMT solver may backjump to decision level 0 and propagate $(\mathsf{cost} \bowtie \mathsf{m})$, further restricting the cost range.

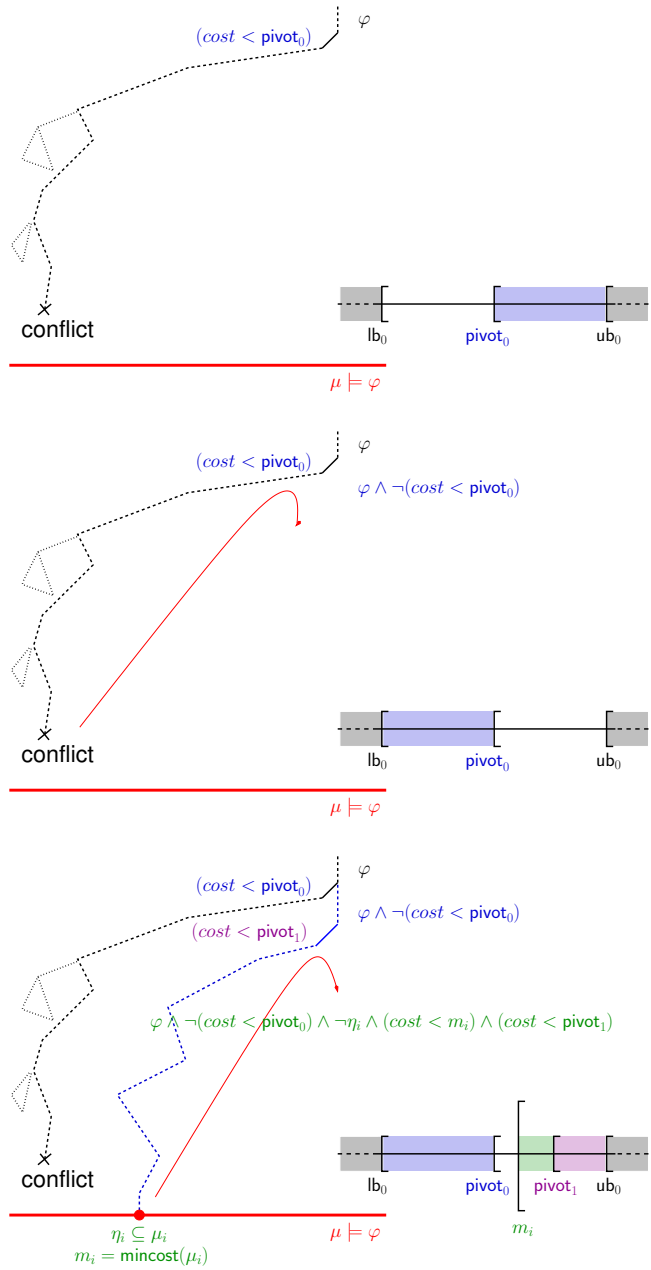

**Fig. 1.** One piece of possible execution of an inline procedure. (i) Pivoting on $(\mathsf{cost} < \mathsf{pivot}_0)$. (ii) Increasing the lower bound to $\mathsf{pivot}_0$. (iii) Decreasing the upper bound to $\mathsf{mincost}(\mu_i)$.

The same considerations about the offline procedure in §4.1 hold for the inline version. The efficiency of the inline procedure can be further improved as follows.

First, when a truth assignment $\mu$ with a novel minimum min is found, not only $(\mathsf{cost} < \mathsf{min})$ but also $\mathsf{PIV} \stackrel{\text{def}}{=} (\mathsf{cost} < \mathsf{pivot})$ is learned as unit clause. Although redundant from the logical perspective because $\mathsf{min} < \mathsf{pivot}$, the unit clause PIV allows the SAT solver for reusing all the clauses in the form $\neg\mathsf{PIV} \vee C$ which have been learned when investigating the cost range $[\mathsf{l}, \mathsf{pivot}[$. (In Algorithm 1 this is done implicitly, since PIV is not popped from $\varphi$ before step 16.) Moreover, the $\mathcal{LA}(\mathbb{Q})$-inconsistent assignment $\mu \wedge (\mathsf{cost} < \mathsf{min})$ may be fed to $\mathcal{LA}(\mathbb{Q})$-Solver and the negation of the returned conflict $\neg\eta \vee \neg(\mathsf{cost} < \mathsf{min})$ s.t. $\eta \subseteq \mu$, can be learned, which prevents the SAT solver from generating any assignment containing $\eta$.

Second, in binary-search mode, if the $\mathcal{LA}(\mathbb{Q})$-Solver returns a conflict set $\eta \cup \{\mathsf{PIV}\}$, then it is further asked to find the maximum value max s.t. $\eta \cup \{(\mathsf{cost} < \mathsf{max})\}$ is also $\mathcal{LA}(\mathbb{Q})$-inconsistent. (This is done with a simple modification of the algorithm in [20].) If $\mathsf{max} \geq \mathsf{u}$, then the clause $C^* \stackrel{\text{def}}{=} \neg\eta \vee \neg(\mathsf{cost} < \mathsf{u})$ is used do drive backjumping and learning instead of $C \stackrel{\text{def}}{=} \neg\eta \vee \neg\mathsf{PIV}$. Since $(\mathsf{cost} < \mathsf{u})$ is permanently assigned at level 0, the dependency of the conflict from PIV is removed. Eventually, instead of using $C$ to drive backjumping to level 0 and propagating $\neg\mathsf{PIV}$, the SMT solver may use $C^*$, then forcing the procedure to stop. If $\mathsf{u} > \mathsf{max} > \mathsf{pivot}$, then the two clauses $C_1 \stackrel{\text{def}}{=} \neg\eta \vee \neg(\mathsf{cost} < \mathsf{max})$ and $C_2 \stackrel{\text{def}}{=} \neg\mathsf{PIV} \vee (\mathsf{cost} < \mathsf{max})$ are used to drive backjumping and learning instead of $C \stackrel{\text{def}}{=} \neg\eta \vee \neg\mathsf{PIV}$. In particular, $C_2$ forces backjumping to level 1 ad propagating the (possibly fresh) atom $(\mathsf{cost} < \mathsf{max})$; eventually, instead of using $C$ do drive backjumping to level 0 and propagating $\neg\mathsf{PIV}$, the SMT solver may use $C_1$ for backjumping to level 0 and propagating $\neg(\mathsf{cost} < \mathsf{max})$, restricting the range to $[\mathsf{max}, \mathsf{u}[$ rather than to $[\mathsf{pivot}, \mathsf{u}[$.

*Example 1.* Consider the formula $\varphi \stackrel{\text{def}}{=} \psi \wedge (\mathsf{cost} \geq a + 15) \wedge (a \geq 0)$ for some $\psi$ in the cost range $[0, 16[$. With basic binary-search, deciding $\mathsf{PIV} \stackrel{\text{def}}{=} (\mathsf{cost} < 8)$, the $\mathcal{LA}(\mathbb{Q})$-Solver produces the $\mathcal{LA}(\mathbb{Q})$-lemma $C \stackrel{\text{def}}{=} \neg(\mathsf{cost} \geq a + 15) \vee \neg(a \geq 0) \vee \neg\mathsf{PIV}$ causing backjumping to level 0 and unit-propagating $\neg\mathsf{PIV}$ on $C$, restricting the range to $[8, 16[$; it takes a sequence of similar steps to progressively restrict the range to $[8, 16[$, $[12, 16[$, $[14, 16[$, and $[15, 16[$. If instead the $\mathcal{LA}(\mathbb{Q})$-Solver produces the $\mathcal{LA}(\mathbb{Q})$-lemmas $C_1 = \neg(\mathsf{cost} \geq a + 15) \vee \neg(a \geq 0) \vee \neg(\mathsf{cost} < 15)$ and $C_2 = \neg\mathsf{PIV} \vee (\mathsf{cost} < 15)$, this first causes backjumping to level 1 the unit-propagation of $(\mathsf{cost} < 15)$ after PIV, and then a backjumping on $C_1$ to level zero, unit-propagating $\neg(\mathsf{cost} < 15)$, which directly restricts the range to $[15, 16[$.

### 4.3 Extensions to OMT($\mathcal{LA}(\mathbb{Q}) \cup \mathcal{T}$)

The procedures of §4.1 and §4.2 extend to the OMT($\mathcal{LA}(\mathbb{Q}) \cup \mathcal{T}$) case straightforwardly as follows. We assume that the underlying SMT solver handles $\mathcal{LA}(\mathbb{Q}) \cup \mathcal{T}$, and that $\varphi$ is a $\mathcal{LA}(\mathbb{Q}) \cup \mathcal{T}$ formula (which for simplicity and wlog we assume to be pure [28]).

We first recall that a $\mathcal{LA}(\mathbb{Q}) \cup \mathcal{T}$ formula $\varphi$ is $\mathcal{LA}(\mathbb{Q}) \cup \mathcal{T}$-satisfiable iff there is a truth assignment

$$\mu \stackrel{\text{def}}{=} \mu_{\mathcal{LA}(\mathbb{Q})} \cup \mu_{ed} \cup \mu_{\mathcal{T}} \qquad (10)$$

where $\mu$ propositionally satisfies $\varphi$, $\mu_{ed}$ is a total truth assignment for the equalities $(x_i = x_j)$ over all the shared variables in $\varphi$ (interface equalities), $\mu_{\mathcal{LA}(\mathbb{Q})} \cup \mu_{ed}$ and $\mu_{\mathcal{T}} \cup \mu_{ed}$ are $\mathcal{LA}(\mathbb{Q})$- and $\mathcal{T}$-satisfiable respectively (see e.g. [28, 15]). (The pedexes $e, d, i$ in $\mu_{...}$ mean "equalities", "disequalities" and "inequalities" respectively.) Thus, the problem reduces to find a truth assignment $\mu$ like (10) s.t. $\mu_{\mathcal{LA}(\mathbb{Q})} \cup \mu_{ed}$ has a $\mathcal{LA}(\mathbb{Q})$-model $\mathcal{M}$ of minimum cost. (Notice that $\mu_{ed}$ is a set of both equalities $(x_i = x_j)$ and disequalities $\neg(x_i = x_j)$.)

Consequently, a naive OMT($\mathcal{LA}(\mathbb{Q}) \cup \mathcal{T}$) variant of the procedures for OMT($\mathcal{LA}(\mathbb{Q})$) of §4.1, §4.2 would be such that the SMT solver enumerates "extended" satisfying truth assignments $\mu$ like (10), checking the $\mathcal{LA}(\mathbb{Q})$- and $\mathcal{T}$-consistency of its $\mu_{\mathcal{LA}(\mathbb{Q})} \cup \mu_{ed}$ and $\mu_{\mathcal{T}} \cup \mu_{ed}$ components and then minimizing the $\mu_{\mathcal{LA}(\mathbb{Q})} \cup \mu_{ed}$ component.

However, minimizing $\mu_{\mathcal{LA}(\mathbb{Q})} \cup \mu_{ed}$ could be computationally problematic due to the presence of disequalities $\neg(x_i = x_j)$s, which would force case-splitting each of them into $(x_i < x_j) \vee (x_j < x_i)$.

A better idea [20, 12] is to let the SAT solver handle this case-splittings, and only when it is necessary. In principle, we could safely augment $\varphi$ with $\mathcal{LA}(\mathbb{Q})$-valid clauses, obtaining

$$\varphi' \stackrel{\text{def}}{=} \varphi \wedge \bigwedge_{x_i, x_j \in SharedVars(\varphi)} ((x_i = x_j) \vee (x_i < x_j) \vee (x_j < x_i)), \qquad (11)$$

which is equivalent to $\varphi$. If we applied the "naive" OMT($\mathcal{LA}(\mathbb{Q}) \cup \mathcal{T}$) procedure above to $\varphi'$, we would generate assignments $\mu$ like (10) in which each $(x_i = x_j)$ would "occur" in either of the following forms:

$$..., (x_i = x_j), \neg(x_i < x_j), \neg(x_j < x_i), ... \qquad (12)$$
$$..., (x_i < x_j), \neg(x_i = x_j), \neg(x_j < x_i), ... \qquad (13)$$
$$..., (x_j < x_i), \neg(x_i = x_j), \neg(x_i < x_j), ... \qquad (14)$$

On the $\mathcal{T}$ side, since $\mathcal{T}$ and $\mathcal{LA}(\mathbb{Q})$ are signature-disjoint, the two strict inequalities are not $\mathcal{T}$-atoms, so that only the equality or disequality is fed to $\mathcal{T}$-Solver. On the $\mathcal{LA}(\mathbb{Q})$ side, since in all the three sub-assignment the first literal entails the other two, the latter ones can be safely dropped from the assignment to be fed to the $\mathcal{LA}(\mathbb{Q})$-solver.

Therefore, overall, an alternative "asymmetric" algorithm works by enumerating truth assignments in the form:

$$\mu' \stackrel{\text{def}}{=} \mu_{\mathcal{LA}(\mathbb{Q})} \cup \mu_{eid} \cup \mu_{\mathcal{T}} \qquad (15)$$

where (i) $\mu'$ propositionally satisfies $\varphi$, (ii) $\mu_{eid}$ is a set of interface equalities $(x_i = x_j)$ and disequalities $\neg(x_i = x_j)$, containing also one inequality in $\{(x_i < x_j), (x_j < x_i)\}$ for every $\neg(x_i = x_j) \in \mu_{eid}$; then $\mu'_{\mathcal{LA}(\mathbb{Q})} \stackrel{\text{def}}{=} \mu_{\mathcal{LA}(\mathbb{Q})} \cup \mu_{ei}$ and $\mu'_{\mathcal{T}} \stackrel{\text{def}}{=} \mu_{\mathcal{T}} \cup \mu_{ed}$ are passed to the $\mathcal{LA}(\mathbb{Q})$-Solver and $\mathcal{T}$-Solver respectively, $\mu_{ei}$ and $\mu_{ed}$ being obtained from $\mu_{eid}$ by dropping the disequalities and inequalities respectively. If $\mu'_{\mathcal{LA}(\mathbb{Q})}$ and $\mu'_\mathcal{T}$ are found consistent in the respective theories, then $\mu'$ is $\mathcal{LA}(\mathbb{Q}) \cup \mathcal{T}$-satisfiable, and so is $\varphi$. Then $\mu'_{\mathcal{LA}(\mathbb{Q})}$ can be fed to Minimize, and hence (cost < min) can be used as new upper bound, as in §4.1 and §4.2.

This motivates and explains the following OMT($\mathcal{LA}(\mathbb{Q}) \cup \mathcal{T}$) variants of the offline and inline procedures of §4.1 and §4.2.

Algorithm 1 is modified as follows. First, SMT.IncrementalSolve in step 8 or 11 is asked to return also a $\mathcal{LA}(\mathbb{Q}) \cup \mathcal{T}$-model $\mathcal{M}$. Then Minimize is invoked on $\langle \text{cost}, \mu_{\mathcal{LA}(\mathbb{Q})} \cup \mu_{ei} \rangle$, s.t. $\mu_{\mathcal{LA}(\mathbb{Q})}$ is the truth assignment over the $\mathcal{LA}(\mathbb{Q})$-atoms in $\varphi$ returned by the solver, and $\mu_{ei}$ is the set of equalities $(x_i = x_j)$ and strict inequalities $(x_i < x_j)$ on the shared variables $x_i$ which are true in $\mathcal{M}$. (The equalities and strict inequalities obtained from the others by the transitivity of $=, <$ can be omitted.)

The implementation of an inline OMT($\mathcal{LA}(\mathbb{Q}) \cup \mathcal{T}$) procedures comes nearly for free if the SMT solver handles $\mathcal{LA}(\mathbb{Q}) \cup \mathcal{T}$-solving by *Delayed Theory Combination* [15], with the strategy of case-splitting automatically disequalities $\neg(x_i = x_j)$ into the two inequalities $(x_i < x_j)$ and $(x_j < x_i)$, which is implemented in MATHSAT. If so the solver enumerates truth assignments in the form $\mu' \stackrel{\text{def}}{=} \mu_{\mathcal{LA}(\mathbb{Q})} \cup \mu_{eid} \cup \mu_\mathcal{T}$, where (i) $\mu'$ propositionally satisfies $\varphi$, (ii) $\mu_{eid}$ is a set of interface equalities $(x_i = x_j)$ and disequalities $\neg(x_i = x_j)$, containing also one inequality in $\{(x_i < x_j), (x_j < x_i)\}$ for every $\neg(x_i = x_j) \in \mu_{eid}$; then $\mu'_{\mathcal{LA}(\mathbb{Q})} \stackrel{\text{def}}{=} \mu_{\mathcal{LA}(\mathbb{Q})} \cup \mu_{ei}$ and $\mu'_\mathcal{T} \stackrel{\text{def}}{=} \mu_\mathcal{T} \cup \mu_{ed}$ are passed to the $\mathcal{LA}(\mathbb{Q})$-Solver and $\mathcal{T}$-Solver respectively, $\mu_{ei}$ and $\mu_{ed}$ being obtained from $\mu_{eid}$ by dropping the disequalities and inequalities respectively. [12]

If this is the case, it suffices to apply Minimize to $\mu'_{\mathcal{LA}(\mathbb{Q})}$, then learn (cost < min) and use it for backjumping, as in §4.2.

## 5 Experimental Evaluation

We have implemented both the OMT($\mathcal{LA}(\mathbb{Q})$) procedures and the inline OMT($\mathcal{LA}(\mathbb{Q}) \cup \mathcal{T}$) procedures of §4 on top of MATHSAT [5] (thus we refer to them as OPT-MATHSAT). We consider four different configurations of OPT-MATHSAT, depending on the approach (offline vs. inline, denoted by "-OF" and "-IN") and the search schema (linear vs. binary, denoted by "-LIN" and "-BIN"). [13] For example, the configuration OPT-MATHSAT-LIN-OF denotes the offline linear-search procedure.

Due to the absence of competitors on OMT($\mathcal{LA}(\mathbb{Q}) \cup \mathcal{T}$), we evaluate the performance of our four configurations of OPT-MATHSAT by comparing them on OMT($\mathcal{LA}(\mathbb{Q})$) problems against GAMS v23.7.1 [16]. GAMS provides two reformulation tools, LOG-MIP v2.0 [4] and JAMS [3] (a new version of the EMP solver [2]), both of them allow

---

[12] In [15] $\mu' \stackrel{\text{def}}{=} \mu_{\mathcal{LA}(\mathbb{Q})} \cup \mu_{ed} \cup \mu_\mathcal{T}$, $\mu_{ed}$ being a truth assignment over the interface equalities, and as such a set of equalities and disequalities. However, since typically a SMT($\mathcal{LA}(\mathbb{Q})$) solver handles disequalities $\neg(x_i = x_j)$ by case-splitting them into $(x_i < x_j) \lor (x_j < x_i)$, the assignment considers also one of the two strict inequalities, which is ignored by the $\mathcal{T}$-Solver and is passed to the $\mathcal{LA}(\mathbb{Q})$-Solver instead of the corresponding disequality.

[13] Here "-LIN" means that BinSearchMode() always returns false, whilst "-BIN" denotes the mixed linear-binary strategy described in §4.1 to ensure termination.

to reformulate LGDP models by using either big-M (BM) or convex-hull (CM) methods
[32, 35]. We use CPLEX v12.2 [23] (through an OSI/CPLEX link) to solve the reformulated MILP models. All the tools were executed using default options, as indicated to
us by the authors [38]. Notice that OPT-MATHSAT uses *infinite precison arithmetic*
whilst, to the best of our knowledge, the GAMS tools implement standard *floating-point arithmetic*.

All tests were executed on 2.66 GHz Xeon machines with 4GB RAM running
Linux, using a timeout of 600 seconds. The correctness of the minimum costs min found
by OPT-MATHSAT have been cross-checked by another SMT solver, either MATHSAT or YICES [7], by detecting the inconsistency within the bounds of $\varphi \wedge (\text{cost} < \text{min})$
and the consistency of $\varphi \wedge (\text{cost} = \text{min})$ (if min is non-strict), or of $\varphi \wedge (\text{cost} \leq \text{min})$ and
$\varphi \wedge (\text{cost} = \text{min} + \epsilon)$ (if min is strict), $\epsilon$ being some very small value. All tools agreed
on the final results, apart from tiny rounding errors, [14] and, much more importantly,
from some noteworthy exceptions on the smt-lib problems (see §5.2).

In order to make the experiments reproducible, the full-size plots, a Linus binary of
OPT-MATHSAT, the problems, and the results are available at [1]. [15]

### 5.1 Comparison on LGDB Problems

We first performed our comparison over two distinct benchmarks, strip-packing and
zero-wait job-shop scheduling problems, which have been previously proposed as benchmarks for LOGMIP and JAMS by their authors [39, 34, 35].

**The strip-packing problem.** Given a set $N$ of rectangles of different length length $L_j$
and height $H_j, j \in 1, .., N$, and a strip of fixed width $W$ but unlimited length, the *strip-packing* problem aims at minimizing the length $L$ of the filled part of the strip while
filling the strip with all rectangles, without any overlap and any rotation. We considered
the LGDP model provided by [34] and a corresponding OMT($\mathcal{LA}(\mathbb{Q})$) encoding.

Every rectangle $j \in J$ is represented by length $L_j$, height $H_j$ and the coordinates
$(x_j, y_j)$ of the upper left corner in the 2-dimensional space. The cost variable to minimize is the length of the strip $L$ which is bounded by the constraints $L \geq (x_i + L_i)$, for
all $i \in I$, and $L \leq \text{ub}$, where ub is an upper bound on the optimal value. For each pair
of rectangles $(i, j)$ where $i, j \in N, i < j$, the encoding contains a disjunction with four
disjuncts $(x_i + L_i \leq x_j), (x_j + L_j \leq x_i), (y_i - H_i \geq y_j)$ and $(y_j - H_j \geq y_i)$ that
constraint their position so that they do not overlap. Three distinct constraints are added
to model the position of each rectangle $j$ in the strip: $(y_i \leq W)$ bounds the $y$-coordinate
by the fixed width of the strip, $(y_i \geq H_i)$ bounds the $y$-coordinate by the height $H_j$ and
$(x_i \leq \text{ub} - L_i)$ bounds the $x$-coordinate.

We randomly generated benchmarks according to a fixed width $W$ of the strip and
a fixed number of rectangles $N$. For each rectangle $j \in N$, length $L_j$ and height $H_j$
are selected in the interval $]0, 1]$ uniformly at random. The upper bound ub is computed with the same heuristic used by [34], which sorts the rectangles in non-increasing order of width and fills the strip by placing each rectangles in the bottom-left corner, and the lower bound lb is set to zero. We generated 100 samples each for 9, 10 and 11 rectangles and for two values of the width $\sqrt{N}/2$ and $1^{16}$.

---

[14] GAMS +CPLEX often gives some errors $\leq 10^{-5}$, which we believe are due to the printing
floating-point format: (e.g. "`3.091250e+00`"); notice that OPT-MATHSAT uses infinite-precision arithmetic, returning values like, e.g. "`7728125177/2500000000`".

[15] We cannot distribute the GAMS tools since they are subject to licencing restrictions. See [16].

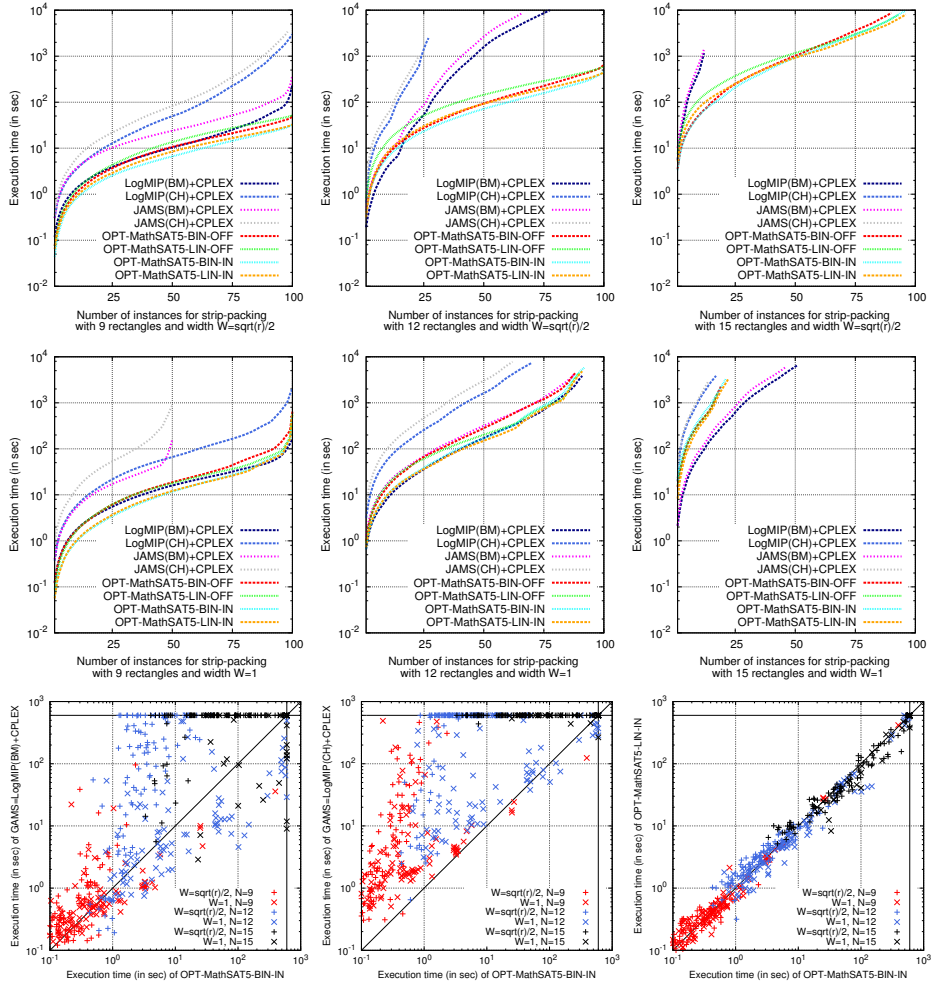

**Fig. 2.** First and second row: cumulative plots of OPT-MATHSAT and GAMS (using LOG-MIP and JAMS) on 100 random instances each of the strip-packing problem for $N$ rectangles, where $N = 9, 12, 15$ (left, center and right respectively), and for width $W = \sqrt{N}/2$ and 1 (first and second row respectively). (The plots for LOGMIP(CH)+CPLEX and JAMS(CH)+CPLEX do not appear since no formula was solved within the timeout.) Third row: comparison of the best configuration of OPT-MATHSAT (OPT-MATHSAT-BIN-IN) against GAMS and OPT-MATHSAT-LIN-IN on 100 random instances of all the strip-packing problem ($N = 9, 12, 15$ and $W = \sqrt{N}/2$ and 1). Left: with LOGMIP(BM)+CPLEX; Center: with LOG-MIP(CH)+CPLEX; Right: with OPT-MATHSAT-LIN-IN.

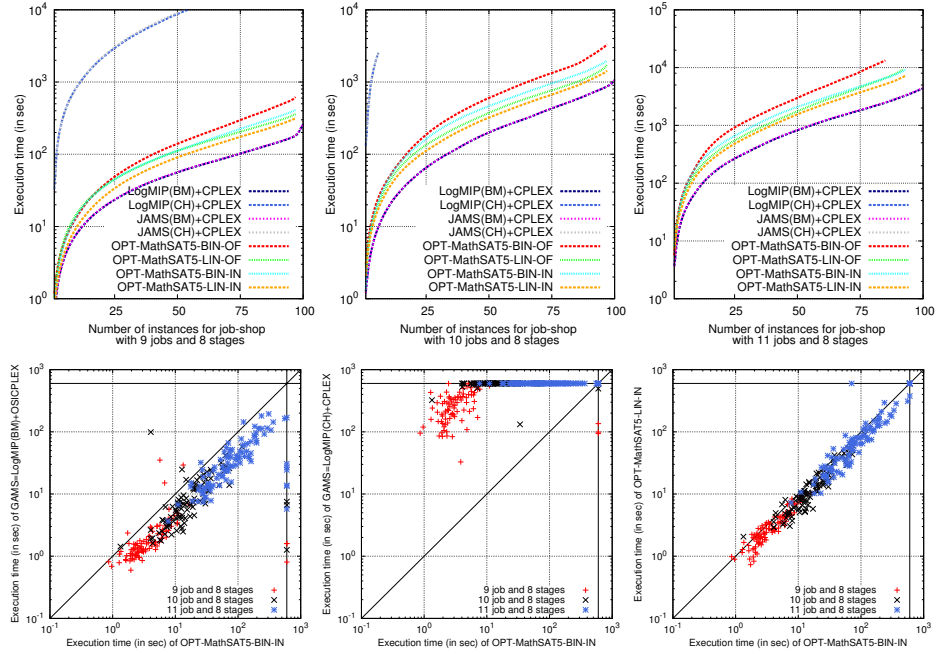

**Fig. 3.** $1^{st}$ row: cumulative plots of OPT-MATHSAT and GAMS on 100 random samples each of the job-shop problem, for $J = 8$ stages and $I = 9, 10, 11$ jobs (left, center and right respectively). Times for LOGMIP(CH)+CPLEX are not reported since no formula was solved within the time-out. $2^{nd}$ row: comparison of the best configuration of OPT-MATHSAT (OPT-MATHSAT-BIN-IN) against GAMS and OPT-MATHSAT-LIN-IN on 100 randomly generated instances of all the job-shop problems ($J = 8, I = 9, 10, 11$). Left: with LOGMIP(BM)+CPLEX; Center: with LOGMIP(CH)+CPLEX; Right: with OPT-MATHSAT-LIN-IN.

The first two rows of the Figure 2 shows the cumulative plots of the execution time for different configurations of OPT-MATHSAT and GAMS on the generated formulas. (The plots for LOGMIP(CH)+CPLEX and JAMS(CH)+CPLEX do not appear since no formula was solved within the timeout.) The third row of Figure 2 compares our best optimization procedure OPT-MATHSAT-BIN-IN against GAMS using LOGMIP with BM and CH reformulation (left and center respectively). The figure also compares our two inline versions OPT-MATHSAT-BIN-IN and OPT-MATHSAT-LIN-IN (right).

**The zero-wait jobshop problem.** Consider the scenario where there is a set $I$ of jobs which must be scheduled sequentially on a set $J$ of consecutive stages with zero-wait transfer between them. Each job $i \in I$ has a start time $s_i$ and a processing time $t_{ij}$ in the stage $j \in J_i$. The goal of the *zero-wait job-shop scheduling* problem is to minimize the makespan, that is the total length of the schedule. In our experiments, we used the LGDP model used in [34] and a corresponding OMT($\mathcal{LA}(\mathbb{Q})$) encoding.

---

[16] Notice that with $W = \sqrt{N}/2$ the filled strip looks approximatively like a square, whilst $W = 1$ is the average of two 2 rectangles.

Every jobs $i \in I$ is represented by the start time $s_i$ and the processing time $t_{ij}$ in stage $j \in J_i$. The cost variable is the makespan $ms$, which is bounded by the constraints ($ms \geq s_i + \sum_{\forall j \in J_i} t_{ij}$), for all $i \in I$. For each pair of jobs $i, k \in I$ and for each stage $j$ with potential clashes (i.e $j \in \{J_i J_k\}$), a disjunction with two disjuncts, ($s_i + \sum_{m \leq j, \forall m \in J_i} t_{im} \leq s_k + \sum_{m < j, \forall m \in J_k} t_{km}$) and ($t_k + \sum_{m \leq j, \forall m \in J_k} t_{km} \leq s_i + \sum_{m < i \forall m \in j_i} t_{im}$), ensures that there is no clash between jobs at any stage at the same time.

We randomly generated benchmarks according to a fixed number of jobs $I$ and a fixed number of stages $J$. For each job $i \in I$, start time $s_i$ and processing time $t_{ij}$ of every job are selected in the interval $]0, 1]$ uniformly at random. We consider a set of 100 samples each for 9, 10 and 11 jobs and 8 stages. We set no value for ub and lb $= 0$.

The first row of Figure 3 shows the cumulative plots of the execution time for different configurations of OPT-MATHSAT and GAMS on the randomly generated formulas. (The plots for LOGMIP(CH)+CPLEX do not appear since no formula was solved within the timeout.) The second row of Figure 3 compares our best optimization procedure OPT-MATHSAT-BIN-IN against GAMS using LOGMIP with BM and CH reformulation (left and center respectively). The figure also compares our two inline versions OPT-MATHSAT-BIN-IN and OPT-MATHSAT-LIN-IN (right).

**Discussion.** Comparing the different version of OPT-MATHSAT, there is no definite winner between -LIN and -BIN options. In fact, OPT-MATHSAT-LIN-OF performs most often better than OPT-MATHSAT-BIN-OF, but OPT-MATHSAT-BIN-IN performances are identical to those of OPT-MATHSAT-LIN-IN. We notice that the -IN options behave uniformly better than the -OF options.

Comparing the different versions of the GAMS tools, we see that (i) on strip-packing instances LOGMIP reformulations lead to better performance than JAMS reformulations, (ii) on job-shop instances they produce substantially identical results. For both reformulation tools, the "BM" versions uniformly outperforms the "CH" ones.

Comparing the different versions of OPT-MATHSAT against all the GAMS tools, we notice that (i) on strip-packing problems all versions of OPT-MATHSAT always outperform all GAMS tools, (ii) on job-shop problems OPT-MATHSAT outperforms the "CM" versions whilst it is beaten by the "BM" ones.

### 5.2 Comparison on SMT-LIB Problems

We compare OPT-MATHSAT against GAMS also on the satisfiable $\mathcal{LA}(\mathbb{Q})$-formulas (QF_LRA) in the SMT-LIB [6]. These instances are all classified as "industrial", because they come from the encoding of different real-world problems in formal verification, planning and optimization. They are divided into six categories: `sc`, `uart`, `sal`, `TM`, `tta_startup`, and `miplib`.[17] Since we have no information on lower bounds on these problems, we use the linear-search version OPT-MATHSAT-LIN-IN. Since we have no control on the origin of each problem and on the name and meaning of the variables, we selected iteratively one variable at random as cost variable, dropping it if the resulting minimum was $-\infty$. This forced us to eliminate a few instances, in particular all `miplib` ones.

---

[17] Notice that other SMT-LIB categories like `spider_benchmarks` and `clock_synchro` do not contain satisfiable instances and are thus not reported here.

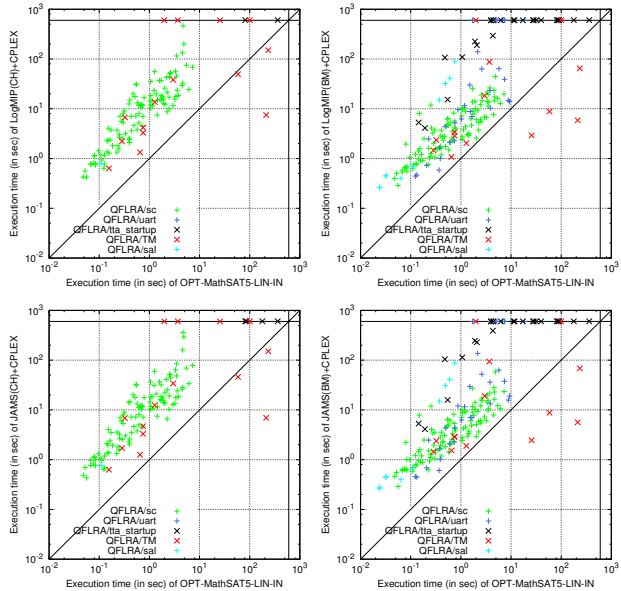

**Fig. 4.** Scatter-plots of the pairwise comparisons on the smt-lib $\mathcal{LA}(\mathbb{Q})$ satisfiable instances between OPT-MATHSAT-BIN-LIN and the two versions of LOGMIP (up) and JAMS. (down).

We first noticed that some results for GAMS have some problem (see Table 1). Using the default options, on $\approx 60$ samples over 193, both GAMS tools with the CH option returned "unfeasible" (inconsistent), whilst the BM ones, when they did not timeout, returned the same minimum value as OPT-MATHSAT. (We recall that all OPT-MATHSAT results were cross-checked, and that the four GAMS tool were fed with the same files.) Moreover, on four `sal` instances the two GAMS tools with BM options returned a wrong minimum value "0", with "CH" they returned "unfeasible", whilst OPT-MATHSAT returned the minimum value "2"; by modifying a couple of parameters from their default value, namely "`eps`" e "`bigM Mvalue`", the results become unfeasible also with BM options.

After eliminating all flawed instances, the results appear as displayed in Figure 4. OPT-MATHSAT solved all problems within the timeout, whilst GAMS did not solve many samples. Moreover, with the exception of 3-4 samples, OPT-MATHSAT always outperforms the GAMS tool, often by more than one order magnitude.

| Solver | Solved | Feasible but ≠ from test | Infeasible | Timeout | Total |
|---|---|---|---|---|---|
| sc | | | | | |
| OPT-MATHSAT-LIN-IN | 108 | 0 | 0 | 0 | 108 |
| JAMS(BM)+CPLEX | 108 | 0 | 0 | 0 | 108 |
| JAMS(CH)+CPLEX | 108 | 0 | 0 | 0 | 108 |
| LOGMIP(BM)+CPLEX | 108 | 0 | 0 | 0 | 108 |
| LOGMIP(CH)+CPLEX | 108 | 0 | 0 | 0 | 108 |
| uart | | | | | |
| OPT-MATHSAT-LIN-IN | 36 | 0 | 0 | 0 | 36 |
| JAMS(BM)+CPLEX | 25 | 0 | 0 | 11 | 36 |
| JAMS(CH)+CPLEX | 0 | 0 | **36** | 0 | 36 |
| LOGMIP(BM)+CPLEX | 25 | 0 | 0 | 11 | 36 |
| LOGMIP(CH)+CPLEX | 0 | 0 | **36** | 0 | 36 |
| tta_startup | | | | | |
| OPT-MATHSAT-LIN-IN | 24 | 0 | 0 | 0 | 24 |
| JAMS(BM)+CPLEX | 8 | 0 | 0 | 16 | 24 |
| JAMS(CH)+CPLEX | 0 | 0 | **20** | 4 | 24 |
| LOGMIP(BM)+CPLEX | 8 | 0 | 0 | 16 | 24 |
| LOGMIP(CH)+CPLEX | 0 | 0 | **22** | 2 | 24 |
| TM | | | | | |
| OPT-MATHSAT-LIN-IN | 15 | 0 | 0 | 0 | 15 |
| JAMS(BM)+CPLEX | 12 | 0 | **1** | 2 | 15 |
| JAMS(CH)+CPLEX | 11 | 0 | 0 | 4 | 15 |
| LOGMIP(BM)+CPLEX | 12 | 0 | **1** | 2 | 15 |
| LOGMIP(CH)+CPLEX | 11 | 0 | 0 | 4 | 15 |
| sal | | | | | |
| OPT-MATHSAT-LIN-IN | 10 | 0 | 0 | 0 | 10 |
| JAMS(BM)+CPLEX | 10 | **4** | 0 | 0 | 10 |
| JAMS(CH)+CPLEX | 1 | 0 | **9** | 0 | 10 |
| LOGMIP(BM)+CPLEX | 10 | **4** | 0 | 0 | 10 |
| LOGMIP(CH)+CPLEX | 1 | 0 | **9** | 0 | 10 |
| check | | | | | |
| OPT-MATHSAT-LIN-IN | 1 | 0 | 0 | 0 | 1 |
| JAMS(BM)+CPLEX | 0 | 0 | **1** | 0 | 1 |
| JAMS(CH)+CPLEX | 0 | 0 | **1** | 0 | 1 |
| LOGMIP(BM)+CPLEX | 0 | 0 | **1** | 0 | 1 |
| LOGMIP(CH)+CPLEX | 0 | 0 | **1** | 0 | 1 |

**Table 1.** Number of solved and correctly-solved problems for OPT-MATHSAT-BIN-LIN and LOGMIP and JAMS CH/BN, on the five groups of smt-lib $\mathcal{LA}(\mathbb{Q})$ satisfiable instances, plus the "chack" instance.

## 6 Conclusions and Future Work

In this paper we have introduced the problem of OMT($\mathcal{LA}(\mathbb{Q}) \cup \mathcal{T}$), an extension of SMT($\mathcal{LA}(\mathbb{Q}) \cup \mathcal{T}$) with minimization of $\mathcal{LA}(\mathbb{Q})$ terms, and proposed two novel procedures addressing it. We have described, implemented and experimentally evaluated this approach, clearly demonstrating all its potentials. We believe that OMT($\mathcal{LA}(\mathbb{Q}) \cup \mathcal{T}$) and its solving procedures propose as a very-promising tools for a variety of optimization problems.

This research opens the possibility for several interesting future directions. A short-term goal is to improve the efficiency and applicability of OPT-MATHSAT: we plan to (i) investigate and implement novel mixed linear/binary-search strategies and heuristics (ii) extend the experimentation to novel sets of problems, possibly investigating ad-hoc customizations. A middle-term goal is to extend the approach to $\mathcal{LA}(\mathbb{Z})$ or mixed $\mathcal{LA}(\mathbb{Q}) \cup \mathcal{LA}(\mathbb{Z})$, by exploiting the solvers which are already present in MATHSAT [22]. A much longer-term goal is to investigate the feasibility of extending the technique to deal with non-linear constraints, possibly using MINLP tools as $\mathcal{T}$-Solver/Minimize.